\documentclass[final]{cvpr}

\usepackage[accsupp]{axessibility}

\usepackage{times}
\usepackage{epsfig}
\usepackage{graphicx}
\usepackage{amsmath}
\usepackage{amssymb}
\usepackage{amsthm}
\usepackage{mathrsfs}
\usepackage{bm}
\usepackage{bbm}
\usepackage{enumitem}
\usepackage{subfigure}
\usepackage{multirow}
\usepackage{url}
\usepackage{pifont}
\newcommand{\cmark}{\checkmark}%
\newcommand{\xmark}{ }%
\newcommand{\ccmark}{\checkmark}%
\newcommand{\xxmark}{\ding{53}}%
\newcommand{\blankM}{\color{white}{\checkmark}}

\usepackage[pagebackref=true,breaklinks=true,colorlinks,bookmarks=false]{hyperref}

\def\cvprPaperID{2487} 
\def\confYear{CVPR 2022}

\newtheorem{prop}{Proposition}

\newcommand{\examplex}{\bm{x}}
\newcommand{\trainingset}{\bm{X}}
\newcommand{\puset}{\bm{X_{PU}}}
\newcommand{\psubset}{\bm{X_{P}}}
\newcommand{\nsubset}{\bm{X_{N}}}
\newcommand{\lsubset}{\bm{X_{L}}}
\newcommand{\usubset}{\bm{X_{U}}}

\newcommand{\ydist}{\mathcal{Y}}

\newcommand{\xdistu}{\mathcal{X_{U}}}

\newcommand{\psudo}{\hat{y}}

\newcommand{\propsudoclass}[1]{{\rm Pr} \left( \hat{y}=#1 | \examplex \right) }

\newcommand{\outputx}{f(\examplex)}

\newcommand{\expectation}[1]{\mathbb{E}_{#1}}
\newcommand{\zeronerror}{\mathbbm{1}\left(\hat{y}, y\right)}

\newcommand{\Ey}{\expectation{(\examplex,y) \sim p(\examplex,y)} \left[y \right]}

\newcommand{\epsdeltntwo}[1]{ \sqrt{ \frac{ \ln{ 4/ \delta} }{2 #1}} }

\newcommand{\bangyan}[1]{#1}
\newcommand{\yunrui}[1]{#1}

\newcommand{\best}[1]{\textcolor[RGB]{205,83,76}{#1}}
\newcommand{\secondb}[1]{\textcolor[RGB]{0,115,194}{#1}}

\begin{document}

\title{Dist-PU: Positive-Unlabeled Learning from a Label Distribution Perspective}

\author{
\parbox{16cm}
  {\centering
    {\large Yunrui Zhao$^{1}$ \ \ \ \ \ \ \ \ \  Qianqian Xu$^{2,*}$ \ \ \ \ \ \ \ \ \  Yangbangyan Jiang$^{3,4}$ \\
    Peisong Wen$^{1,2}$ \ \ \ \ \ \ \ \ \ \ \  Qingming Huang$^{1,2,5,*}$ }\\
    {\normalsize
    $^1$ School of Computer Science and Technology, University of Chinese Academy of Sciences \\
    $^2$ Key Laboratory of Intelligent Information Processing, Institute of Computing Technology, CAS\\
    $^3$ State Key Laboratory of Information Security, Institute of Information Engineering, CAS\\
    $^4$ School of Cyber Security, University of Chinese Academy of Sciences\\
    $^5$ Key Laboratory of Big Data Mining and Knowledge Management, University of Chinese Academy of Sciences\\
    }
    {\tt\small zhaoyunrui20@mails.ucas.ac.cn \quad\quad \{xuqianqian,wenpeisong20z\}@ict.ac.cn\quad\quad jiangyangbangyan@iie.ac.cn\quad\quad qmhuang@ucas.ac.cn  
    }
  }
}

\maketitle

\begin{abstract}
Positive-Unlabeled (PU) learning tries to learn binary classifiers from a few labeled positive examples with many unlabeled ones. \bangyan{Compared with ordinary semi-supervised learning, this task is much more challenging due to the absence of any known negative labels.} While existing cost-sensitive-based methods have achieved state-of-the-art performances, they explicitly \bangyan{minimize the risk of classifying unlabeled data as negative samples}, which might result in a negative-prediction preference of the classifier. \yunrui{To alleviate this issue, we resort to a label distribution perspective for PU learning in this paper.} \bangyan{Noticing that the label distribution of unlabeled data is fixed when the class prior is known, it can be naturally used as learning supervision for the model. Motivated by this, we propose to pursue the label distribution consistency between predicted and ground-truth label distributions, which is formulated by aligning their expectations.} Moreover, we further adopt the entropy minimization and Mixup regularization to avoid the trivial solution of the label distribution consistency on unlabeled data and mitigate the consequent confirmation bias. Experiments on \bangyan{three benchmark datasets} validate the effectiveness of the proposed method. 
Code available at: \href{https://github.com/Ray-rui/Dist-PU-Positive-Unlabeled-Learning-from-a-Label-Distribution-Perspective}{https://github.com/Ray-rui/Dist-PU-Positive-Unlabeled-Learning-from-a-Label-Distribution-Perspective}.
\end{abstract}

\section{Introduction}
With the advent of big data, deep neural networks have attracted extensive attention, since their performance has reached or even surpassed the human level in various tasks \yunrui{\cite{DBLP:journals/corr/abs-1712-00409,DBLP:journals/jmlr/RaffelSRLNMZLL20,DBLP:conf/eccv/MahajanGRHPLBM18}}. Specifically, such great success usually relies on \bangyan{supervision by a large amount of labeled data}. However, it is hard to obtain intact label information \bangyan{in many real-world applications, even those with only binary options}. For example, observed interactions between users and items in recommendation systems are labeled positives. Since many unconcerned factors like lack of exposure or other coincidences could account for missing interactions, we cannot view all the unobserved interactions as negatives. Similar scenarios include Alzheimer's disease recognition \cite{DBLP:conf/icml/ChenCCYG0W20}, malicious URL detection \cite{DBLP:conf/ccs/ZhangLZLLZZ17} and particle picking in cryo-electron micrographs \cite{bepler2019positive}, where we only have access to a few labeled positives with plenty of unlabeled data. \yunrui{Such a great demand motivates us to learn from positive and unlabeled data, also known as PU learning.}

\begin{figure}[t]
\begin{center}
    \subfigure{
    \centering
    \begin{minipage}[t]{0.225\textwidth}
        \includegraphics[width=1.5in,trim=2 2 2 2,clip]{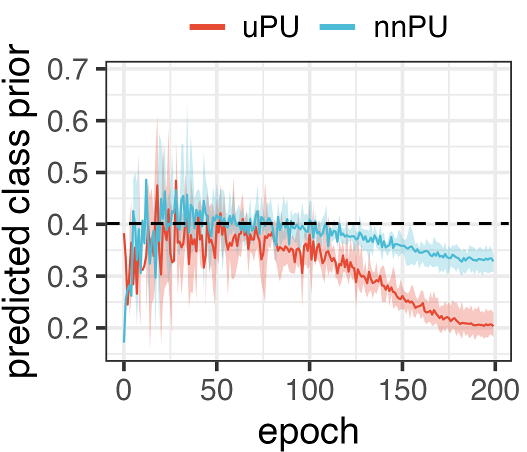}
    \end{minipage}
    }
    \subfigure{
    \centering
    \begin{minipage}[t]{0.225\textwidth}
        \includegraphics[width=1.5in,trim=2 2 2 2,clip]{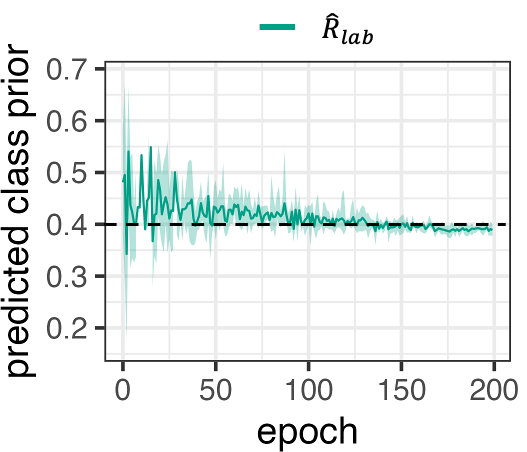}
    \end{minipage}
    }
\end{center}
   \caption{Predicted label distributions of uPU \cite{DBLP:conf/nips/PlessisNS14}, nnPU \cite{DBLP:conf/nips/KiryoNPS17}, and Ours on the training data of CIFAR-10 with 5 repeats and Vehicles as the positive class. Compared with uPU and nnPU, the proposed label distribution alignment scheme ($\hat{R}_{lab}$) rectifies the negative-prediction preference, and achieves a predicted class prior consistent with the ground-truth (black dashed lines).}
\label{fig:motivation}
\end{figure}

\yunrui{Researchers} have developed numerous PU algorithms over the past decades. 
One prevalent \yunrui{research line} would be cost-sensitive PU learning. These methods \cite{DBLP:conf/nips/PlessisNS14,DBLP:conf/icml/PlessisNS15,DBLP:conf/nips/KiryoNPS17,DBLP:conf/ijcai/SuCX21} often assume the availability of class prior and \bangyan{minimize the risk of classifying} unlabeled data as negative \yunrui{instances}. By reweighting the importance of the positive and the negative risks, they could get an either unbiased or consistent risk estimator for PU learning. 
\bangyan{Despite the great success the cost-sensitive methods have achieved, explicitly optimizing the risk that classifies unlabeled data to the negative class would essentially lead a flexible model (\eg, deep neural networks) to overfit, and further result in a negative-prediction preference of the classifier, as illustrated in the left half of Fig.~\ref{fig:motivation}. As the training epoch increases, the predicted class prior probability ${\rm Pr}(\hat{y}=1)$, \ie, the expectation of predicted label distribution, tends to decrease from the ground-truth class prior.} \bangyan{On the other hand, we notice that given the class prior, the underlying label distribution of data is immediately determined. Such a label distribution could be a natural supervision for the model. More specifically,} \yunrui{the distribution of the model's predicted labels is supposed to be consistent with that of ground-truth ones. Following this intuition, we propose the label distribution alignment for PU learning, which aligns the expectations of the predicted and the ground-truth labels to ensure the label distribution consistency.} In particular, the expectation of predicted labels is estimated by sigmoid outputs from a deep network, \bangyan{enabling the end-to-end learning of the proposed framework. Compared with the cost-sensitive methods, the proposed label distribution alignment scheme could rectify the negative-prediction preference (see Fig.~\ref{fig:motivation}).} 

\bangyan{Nevertheless, merely pursuing the label distribution consistency might suffer from} \yunrui{a trivial solution that all the predicted scores of unlabeled data are equal to the class prior.} To avoid this issue, we employ the entropy minimization technique, which encourages the model to produce scores much closer to zero or one. Meanwhile, Mixup \cite{DBLP:conf/iclr/ZhangCDL18} is further adopted to alleviate the confirmation bias caused by the model's overfitting on its early predictions. To summarize, the contributions of this paper are three-fold:
\begin{itemize}[leftmargin=*]
    \setlength{\itemsep}{0pt}
    \setlength{\parsep}{0pt}
    \setlength{\parskip}{0pt}
    \item We propose a PU learning framework based on label distribution alignment called \textbf{Dist-PU}. Different with existing methods, \textbf{Dist-PU} aligns the expectation of the model's predicted labels with that of ground-truth ones, and thus mitigates the negative-prediction preference. We also present a generalization bound for label distribution alignment as the theoretical guarantee.
    \item We further incorporate entropy minimization and Mixup to avoid the trivial solution of label distribution alignment and the consequent confirmation bias.
    \item Extensive experiments are conducted on Fashion-MNIST, CIFAR-10 and Alzheimer datasets, where \textbf{Dist-PU} outperforms existing state-of-the-art models in most cases.
\end{itemize}


\section{Related work}
PU learning models mainly fall into two categories \cite{DBLP:journals/ml/BekkerD20}. One prevalent research line adopts the framework of cost-sensitive learning. During the optimization step, samples relate to different importance weights. The unbiased risk estimator of PU learning, known as uPU, was proposed by Plessis, Niu, and Sugiyama \cite{DBLP:conf/nips/PlessisNS14}. Later, the authors of uPU found that a convex surrogate loss could reduce the computational cost \cite{DBLP:conf/icml/PlessisNS15}. Since then, works have surged to enhance this technique. Due to the deep models’ strong fitting ability, the empirical risk of training data could go negative. Therefore, the non-negative risk estimator, known as nnPU, was proposed by Kiryo \cite{DBLP:conf/nips/KiryoNPS17}. Besides, Self-PU \cite{DBLP:conf/icml/ChenCCYG0W20} introduces self-supervision to nnPU via auxiliary tasks \yunrui{including} model calibration and distillation \yunrui{with a self-paced curriculum}; ImbPU \cite{DBLP:conf/ijcai/SuCX21} extends nnPU to imbalanced data \yunrui{by magnifying the weights of minority class.} However, all the above methods assume identical distributions between labeled positives and ground-truth ones. PUSB \cite{DBLP:conf/iclr/KatoTH19} relaxes such an assumption by maintaining the order-preserving property. Furthermore, aPU \cite{DBLP:conf/nips/HammoudehL20} deals with an arbitrary positive shift between source and target distributions.

Another branch of PU learning employs two heuristic steps. Such methods \cite{DBLP:conf/icml/LiuLYL02,DBLP:journals/tkde/YuHC04} 
first identify reliable negative or positive examples from the unlabeled data, thus yielding (semi-) supervised learning in the second step. The two-step methods differ in ways of \yunrui{assigning labels to unlabeled data}. 
Graph-based methods \cite{DBLP:conf/nips/ZhouBLWS03,DBLP:conf/iconip/ChaudhariS12,DBLP:journals/jcp/ZhangZ09} measure distances between samples through graphs to affirm the labels of the unlabeled data. RP \cite{northcutt2017rankpruning} learns from confident examples whose predicted scores are near zero or one. PUbN \cite{DBLP:conf/icml/HsiehNS19} pretrains a model with nnPU to recognize some negatives. It then combines the positive risk, the unlabeled risk, and the negative risk to learn the final classifier. GenPU \cite{hou2018generative} is established from a generative learning perspective. It leverages the GAN framework and uses the generated data to train the final classifier. KLDCE\cite{gong2019loss} firstly translates PU learning into a label noise problem and weakens its side effect secondly via centroid estimation of the corrupted negative set. PULNS \cite{DBLP:conf/aaai/LuoZCQDZWCHRL21} incorporates reinforcement learning to obtain an effective negative sample selector.

Most PU learning methods are established on a known class prior. To this end, there emerge some class prior estimation algorithms. PE \cite{DBLP:conf/acml/PlessisNS15} uses partial matching and minimizes the Pearson divergence between the unlabeled and the labeled distributions. Pen-L1  \cite{DBLP:journals/ml/PlessisNS17} corrects the overestimate of PE. KM1 and KM2  \cite{DBLP:conf/icml/RamaswamyST16} model the positive distribution with the distance between kernel embeddings. TIcE \cite{DBLP:conf/aaai/BekkerD18} 
approximates the class prior through a decision tree induction. CAPU \cite{DBLP:conf/ijcai/Chang0Z20} and $(TED)^n$ \cite{DBLP:conf/nips/GargWSBL21} estimate the class prior and learn a classifier jointly. Other works pursue for PU learning without class prior. VPU \cite{DBLP:conf/nips/ChenLWZW20} proposes a variational principle for PU learning with Mixup regularization. PAN \cite{DBLP:conf/aaai/HuL0JM0021} revises the architecture of GAN and proposes a new objective based on KL-distance. Instance-dependent PU learning \cite{gong2021instance} treats the class label as a hidden variable, aquiring the classifier under the EM framework.

\section{Methodology}


\subsection{Problem setting}
\bangyan{In binary classification, the input space is $\mathcal{X} \subseteq \mathbb{R}^{d}$ with $d$ dimensions, and the label space is $\ydist = \left\{0,1\right\}$ with $0$ for negative and $1$ for positive. Let $p(\examplex, y)$ be the underlying probability density of $(\mathcal{X},\ydist)$, and $p(\examplex)$ be the marginal distribution of the input. Then the sets of positive and negative samples could be denoted as:
\begin{equation}
    \begin{aligned}
    \psubset = & \left\{ \examplex \right\}^{n_{P}} \sim p_{P}\left(\examplex\right), \\
    \nsubset = & \left\{ \examplex \right\}^{n_{N}} \sim p_{N}\left(\examplex\right),
    \end{aligned}
\end{equation}
where $p_{P}\left(\examplex\right)$ and $p_{N}\left(\examplex\right)$ denote the class-conditional distribution of positive and negative samples respectively. Based on this, the whole sample set $\trainingset = \psubset \cup \nsubset$ is formulated as:
\begin{align}
    \trainingset &= \left\{ \examplex \right\}^{n} \sim \ p\left(\examplex\right), \\
    p\left(\examplex\right) &= \pi_{P} \cdot p_{P}\left(\examplex\right) + \pi_{N} \cdot p_{N}\left(\examplex\right),
    \label{eq:p_n_u}
\end{align}
where $n=n_{P}+n_{N}$, $\pi_{P}={\rm Pr}(y=1)$ is the class prior probability and $\pi_{N}=1-\pi_{P}$.}

We aim to learn a function $f \in \mathcal{F}: \mathbb{R}^{d} \rightarrow \mathbb{R}$ to minimize the average prediction error, also known as the expected risk:
\begin{align}
    R = \expectation{(\examplex,y) \sim p(\examplex,y)} \left[ \zeronerror \right], \label{eq:exp_risk}
\end{align}
with $\hat{y}=f(\examplex) \in \left\{ 0,1 \right\}$ denoting the predicted label of $\examplex$ by $f$; $\zeronerror$ refers to the zero-one error, which equals $0$ when $\hat{y}=y$, and otherwise $1$. Unfortunately,  $\zeronerror$ is discontinuous, making the model hard to optimize. Therefore, a surrogate loss $l(\cdot,\cdot)$ defined on $f(\examplex)$ and $y$ is usually used for risk minimization.

PU learning is a special case of binary classification, in the way that only a small portion of positive data are labeled. Formally, the training set $\puset = \lsubset \cup \usubset$ where $\lsubset$ or $\usubset$ represents the labeled positive or the unlabeled subset respectively. Under the Selected Completely at Random (SCAR) assumption and the case-control scenario \bangyan{\cite{DBLP:journals/ml/BekkerD20}}, positives are labeled uniformly at random and independently of their features, while the unlabeled data are i.i.d drawn from the real marginal distribution:
\begin{align}
    \lsubset = & \left\{ \examplex \right\}^{n_{L}} \sim  p_{P}\left(\examplex\right), 
    \label{eq:dist_lp} \\
    \usubset = & \left\{ \examplex \right\}^{n_{U}} \sim  p_{U}\left(\examplex\right)=p\left(\examplex\right). \label{eq:dist_u}
\end{align}

\subsection{Label distribution alignment}

\begin{figure*}
\begin{center}
\includegraphics[width=1\linewidth,trim= 4 8 4 4,clip]{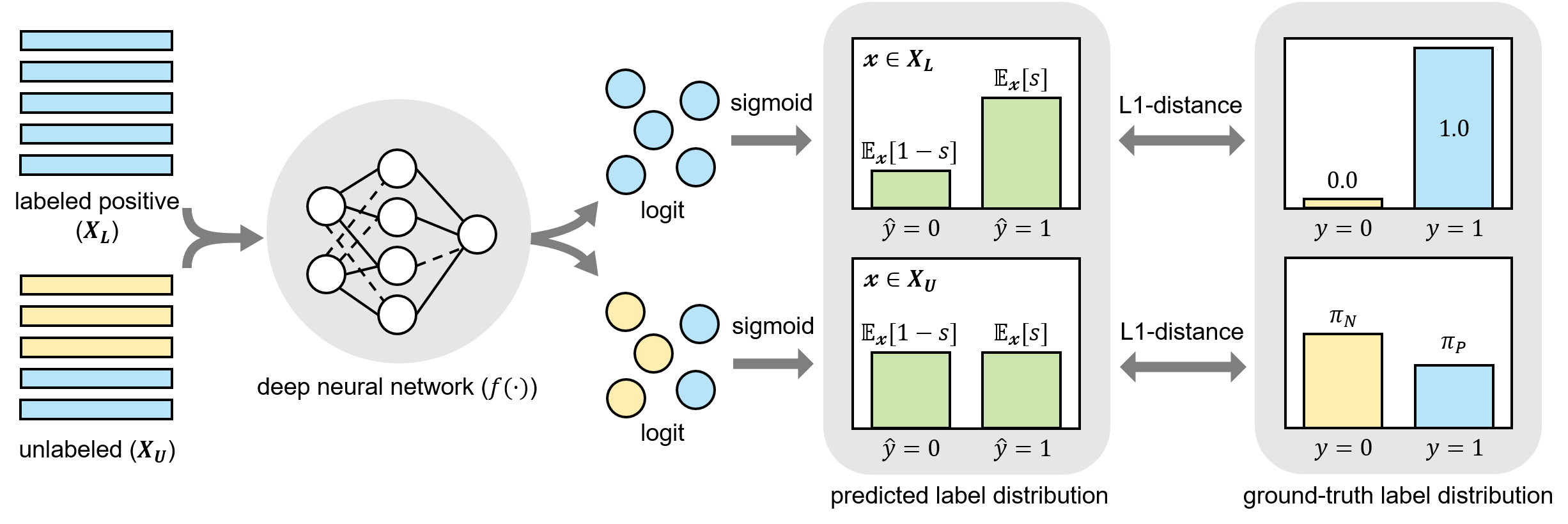}
\end{center}
   \caption{Overview of the proposed label distribution alignment framework. \bangyan{We aim at matching the predicted label distributions over labeled positive and unlabeled data with their ground-truth distributions by aligning the corresponding expectations.} Specifically, a labeled positive subset ($\lsubset$) and unlabeled data ($\usubset$) are \bangyan{fed into} a deep neural network ($f$) \bangyan{followed by a sigmoid function. Using the output score $s$, we could approximately estimate the expectation of $\hat{y}$ by $\mathbb{E}_{\examplex} [s]$.} \yunrui{Note that the expectation of the ground-truth label $y$'s distribution of $\lsubset$ is $1$, while that of $\usubset$ is equal to the class prior $\pi_{P}$. Then, we train $f$ by minimizing the L1-distances between the expectations of the predicted and the ground-truth labels from $\lsubset$ and $\usubset$, respectively.}}
\label{fig:overview}
\end{figure*}

The key to PU learning lies in the ways of incorporating unlabeled data $\usubset$ into the training process. \bangyan{Without knowing any label in $\usubset$, we cannot directly calculate a loss like $\ell\left( f(\examplex), y\right)$. As a solution, a large proportion of existing methods rely on the risk of classifying unlabeled data as positive or negative samples. Yet an essential fact might be ignored that the expectation of all the ground-truth labels over the entire data distribution is exactly the class prior, \ie, $\Ey = \pi_{P}$. Motivated by this, we propose to use the class prior $\pi_{P}$ to guide the learning of the model. More specifically, we pursue} the consistency between the predicted and ground-truth label distributions of labeled and unlabeled data, offering relatively reliable supervision. \bangyan{As shown in Fig.~\ref{fig:overview}, for labeled positives $\lsubset$, the expectation of predicted labels should be equal to 1. Meanwhile, in $\usubset$, positives are expected to take over $\pi_{P}$ proportion, and $\pi_{N}$ for negatives. Thus, the expectation over $\usubset$ should be matched to $\pi_{P}$. By aligning the expectations of predicted and ground-truth label distributions, the label distribution consistency could be achieved.} Such a mechanism is called as \textit{label distribution alignment} in this paper.

In order to infer the formulation of the label distribution alignment, we first reformulate the expected risk $R$ of Eq.~(\ref{eq:exp_risk}) as a combination of the expected risks on positive and negative classes:
\begin{equation}
    R =  \pi_{P} \expectation{\examplex \sim p_{P}(\examplex)} \left[ \mathbbm{1}(\hat{y},1) \right] 
       + \pi_{N} \expectation{\examplex \sim p_{N}(\examplex)} \left[ \mathbbm{l}(\hat{y},0) \right]. \label{eq:risk2part}
\end{equation}
\bangyan{Since $\hat{y},y \in \left\{ 0,1 \right\}$, we have  $\zeronerror = \left| \hat{y}-y \right| $. Moreover, for  all the positive samples, we can rewrite its loss by $\zeronerror=1-\hat{y}$. Likewise, the zero-one loss of negative samples could be written as $\zeronerror=\hat{y}$.} Then $R$ in Eq.(\ref{eq:risk2part}) \bangyan{could be reformulated as}:
\begin{align}
     R = & \pi_{P} \expectation{\examplex \sim p_{P}(\examplex)} \left[ 1-\hat{y} \right] + \pi_{N} \expectation{\examplex \sim p_{N}(\examplex) } \left[ \hat{y} \right], \nonumber \\
    = & \pi_{P} \underbrace{\left| \expectation{\examplex \sim p_{P}(\examplex)} \left[ \hat{y} \right] - 1 \right|}_{R_{P}} + \pi_{N} \underbrace{\left| \expectation{\examplex \sim p_{N}(\examplex)} \left[ \hat{y} \right] \right|}_{R_{N}}, \label{eq:rprn}
\end{align}
where $R_{P}$ and $R_{N}$ are the expected risk of \bangyan{positive and negative samples,} respectively. \bangyan{Such a reformulation naturally induces} the principle of the label distribution consistency we pursue, \ie, the expectation of $\hat{y}$ should be equal to that of $y$. In other words, given a dataset, the class probabilities of the predicted labels should be consistent with the \bangyan{ground-truth class priors}.

\bangyan{With this principle in mind, we need to find a way to estimate $R_P$ and $R_N$ in PU learning.} In this setting, $R_{P}$ could be evaluated directly from the labeled positive samples, while the situation is not the same for $R_{N}$ due to the absence of any known negative labels. \bangyan{Instead, we may resort to} the label distribution consistency over unlabeled data. According to Eq.~(\ref{eq:dist_u}), \yunrui{the distribution of the unlabeled data is the same with the real distribution. Therefore,} we first concentrate on the difference between the expectation of the predicted labels and that of the underlying ground-truth labels \yunrui{concerning the real distribution}:
\begin{align}
    & \expectation{\examplex \sim p_U(\examplex)} \left[ \hat{y} \right] - \expectation{(\examplex,y) \sim p(\examplex,y)}  \left[ y \right] \nonumber \\ = & \pi_{P}  \expectation{\examplex \sim p_{P}(\examplex)} \left[ \hat{y} \right] - \pi_{P} \nonumber + \pi_{N} \expectation{\examplex \sim p_{N}(\examplex)} \left[ \hat{y} \right] \nonumber \\
        = &  -\pi_{P} R_{P} + \pi_{N} R_{N}.
\end{align}
Then we could \bangyan{easily represent $R_{N}$ by $R_{P}$ and the expectations of label distributions over unlabeled data}:
\begin{align}
    \pi_{N} R_{N} = & \pi_{P} R_{P} + \expectation{\examplex \sim p_U(\examplex)} \left[ \hat{y} \right] - \expectation{(\examplex,y) \sim p(\examplex,y)} \left[ y \right] \nonumber \\ 
    = & \pi_{P} R_{P} + \expectation{\examplex \sim p_U(\examplex)} \left[ \hat{y} \right] - \pi_{P} . \label{eq:rn_label}
\end{align}
\bangyan{By substituting Eq.~(\ref{eq:rprn}) into Eq.~(\ref{eq:rn_label}), we could derive an} equivalent form of $R$ in a label distribution alignment manner, \bangyan{which does not involve any explicit calculation using negative labels}:
\begin{align}
    R = 2\pi_{P} R_{P} + \expectation{\examplex \sim p_U(\examplex)} \left[ \hat{y} \right] - \pi_{P}.
\end{align}
However, \yunrui{due to} insufficient label information and weak supervision for unlabeled data, the expectation of the predicted labels \yunrui{might be unstable during the training process}. It could deviate far from the ground-truth prior, \yunrui{even making the difference of $\expectation{\examplex \sim p(\examplex)} \left[ \hat{y} \right]$ and $\pi_{P}$ negative.} \bangyan{In this case, continuing minimizing this risk might worsen the situation. On the other hand, taking the label distribution consistency into account,} we \bangyan{wish} that $\expectation{\examplex \sim p(\examplex)} \left[ \hat{y} \right]$ is \yunrui{exactly} $\pi_{P}$. Therefore, we propose a variant risk by putting an absolute function into $R$:
\begin{equation}
    R_{lab} = 2\pi_{P} R_{P} + \underbrace{ \left| \expectation{\examplex \sim p_U(\examplex)} \left[ \hat{y} \right] - \pi_{P} \right|}_{R_{U}}.
\end{equation}

Though by this equation we decompose the risk \bangyan{over the entire data distribution} into terms which can be estimated using \bangyan{labeled positive and unlabeled samples}, there still exist some practical issues \bangyan{when applying the Empirical Risk Minimization (ERM) principle \cite{DBLP:books/daglib/0097035} for its optimization, \ie, the differentiability of $\psudo$}. It is common and straightforward to determine the predicted label via the sign of the model's output score $\outputx$. \bangyan{Namely, $\psudo=\frac{1}{2}(\text{sgn}(\outputx)+1)$. This process suffers from the non-differentiability of the sign function. As a result, the induced empirical objective could not be updated 
in an end-to-end manner. To mitigate this issue,} we turn to directly consider $\mathbb{E}_{\examplex} \left[ \hat{y} \right]$ and compute it by the conditional probability of $\hat{y}$ given $\examplex$:
\begin{align}
    \mathbb{E}_{\examplex} \left[ \hat{y} \right] = \mathbb{E}_{\examplex} \left[ \propsudoclass{1} \cdot 1 + \propsudoclass{0} \cdot 0 \right].
\end{align}
The problem then shifts to how to aquire the conditional probability. Through a sigmoid function, we can approximately calculate $\propsudoclass{1}$ by $s$ using $\outputx$ as follows:
\begin{align}
    s = \frac{1}{1+\exp{\left[ -\outputx \right]}}.
\end{align}
In this manner, \bangyan{we could alternatively employ the differentiable $\mathbb{E}_{\examplex} \left[ s \right]$ instead of the original expectation of hard predicted labels}, furtherly making the optimization compatible with a gradient descent style.

To sum up, we obtain the optimization objective of our label distribution alignment as follows:
\begin{align}
    \hat{R}_{lab} = 2\pi_{P} \underbrace{\left| \frac{1}{n_{L}} \sum_{\examplex \in \lsubset} s - 1 \right|}_{\hat{R}_{L}} + \underbrace{\left| \frac{1}{n_{U}} \sum_{\examplex \in \usubset} s - \pi_{P} \right|}_{\hat{R}_{U}}, \label{eq:distpurisk}
\end{align}
where $\hat{R}_{L}$ and $\hat{R}_{U}$ denote the empirical risk estimator of $R_{P}$ and $R_{U}$, respectively.

\bangyan{Though the label distribution alignment risk $\hat{R}_{lab}$ is obviously biased, the following proposition shows that the original risk $R$} can be upper bounded by $\hat{R}_{lab}$ together with sample sizes and model complexity based terms.
\begin{prop}
For a class of $b$-uniformly bounded functions $\mathcal{F}$ with the VC dimension $\mathcal{V}$, with a probability at least $1-\delta$, it holds that:
\begin{align}
    R \leq & 2 \hat{R}_{lab} + 8\pi_{P} \cdot C \sqrt{\frac{\mathcal{V}}{n_{L}}} + 12\pi_{P} \epsdeltntwo{n_{L}} \nonumber \\
    & + 4 C \sqrt{\frac{\mathcal{V}}{n_{U}}} + 6 \epsdeltntwo{n_{U}},
\end{align}
where $C$ is a universal constant. 
\end{prop}
Therefore, optimizing the upper bound naturally leads to a minimization of $R$. Since all the terms except $\hat{R}_{lab}$ only depend on the dataset sizes and the complexity of the model, here we could minimize $\hat{R}_{lab}$ to indirectly optimize $R$. Moreover, we also see that using more labeled positive or unlabeled samples, or a less-complex model (\ie, smaller $\mathcal{V}$) would reduce the generalization gap. Proof is provided in supplementary materials due to the space limitation.


\subsection{Entropy minimization}
So far, Eq.~(\ref{eq:distpurisk}) seems to serve our pursuit well. However, we would like to remind that Eq.~(\ref{eq:distpurisk}) might induce a trivial solution. Consider the trivial solution that every $s$ is caculated as $\pi_{P}$ regradless of the inputs, $\hat{R}_{U}$ will reach $0$ as well, which is of course against our hope. We hope that on the premise of label distribution consistency, $s$ is as close to either $0$ or $1$ as possible. To this aim, we introduce the \textit{entropy minimization} technique to concentrate the probability distribution on a single class, thus avoiding the trivial solution. Entropy minimization appears widely in semi-supervised learning \cite{DBLP:conf/nips/GrandvaletB04,DBLP:conf/nips/XieDHL020,DBLP:conf/nips/BerthelotCGPOR19,DBLP:conf/iclr/BerthelotCCKSZR20}. It originates from the well-known clustering hypothesis that the data near the hyperplane of the classification decision is sparse, and the data in the place of the same category cluster is dense. According to this hypothesis, we expect low entropy of the scores obtained from the unlabeled data: 
\begin{equation}
    L_{ent} = -\frac{1}{n_{U}} \sum_{\examplex \in \usubset} \left[ (1-s) \log{(1-s)} + s \log{s} \right]. \label{eq:ent}
\end{equation}

\subsection{Confirmation bias}
Although \bangyan{the incorporation of entropy minimization could result in a sharper bimodal distribution for predicted scores of unlabeled data, it might also further increase over-confidence of the model. Such over-confidence is especially fatal during the early training stage.} Incorrect predictions might get reinforced as the training process steps forward, while the label distribution consistency and entropy minimization make the situation even worse. \bangyan{The phenomenon of} overfitting wrongly predicted data, known as \textit{confirmation bias}, is \bangyan{ubiquitous} in semi-supervised learning \cite{DBLP:conf/ijcnn/ArazoOAOM20, DBLP:conf/iclr/LiSH20}. \bangyan{In this paper, we resort to the popular Mixup regularization \cite{DBLP:conf/iclr/ZhangCDL18} to deal with the confirmation bias.}

Mixup is motivated by the Vicinal Risk Minimization (VRM) principle \cite{DBLP:conf/nips/ChapelleWBV00}. It randomly combines training pairs $\bm{x_{1}}, \bm{x_{2}} \in \puset$ convexly \bangyan{with the mixing proportion $\lambda$ chosen from a Beta distribution}:
\begin{align}
    \lambda \sim & \rm{Beta}(\alpha, \alpha), \label{eq:alpha} \\
    \lambda' = & \max(\lambda, 1-\lambda), \label{eq:lambda_mix} \\ 
    \bm{x'} = & \lambda'\bm{x_{1}}+(1-\lambda')\bm{x_{2}}. \label{eq:x_mix}
\end{align}
Then the regularization term leads the mixed sample $\bm{x'}$ to have a consistent prediction with the linear interpolation of the associated targets:
\begin{align}
    L_{mix} = \frac{1}{n} \sum_{\bm{x_{1}},\bm{x_{2}} \in \puset} & [ \lambda' l_{bce}(s',s_{1}) \nonumber \\
    & + (1-\lambda') l_{bce}(s',s_{2}) ], \label{eq:l_mix}
\end{align}
where $l_{bce}(t',t)=-(1-t)\log{(1-t')}-t\log{(t')}$ is the binary cross entropy loss; $s'$ is the predicted score of $\bm{x'}$, while $s_{1} (s_{2})$ refers to the corresponding soft label (\ie, predicted score) of $\bm{x_{1}} (\bm{x_{2}})$. 

The reasons why Mixup can alleviate confirmation bias are three-folds. Firstly, using predicted scores as soft labels in Eq.~(\ref{eq:l_mix}) reduces the model's over-confidence, balancing the trade-off between entropy minimization and confirmation bias. Secondly, Mixup loss is more robust to prediction errors. For instance, considering two examples \bangyan{that are predicted as false negative and false positive, respectively}, their Mixup loss will still be helpful especially when $\lambda'$ in Eq.~(\ref{eq:lambda_mix}) is near $0.5$. Thirdly, Mixup \bangyan{could be regarded as a kind of data augmentation \cite{DBLP:conf/nips/SimardLDV96}}, which extends the training dataset by additional virtual examples as shown in Eq.~(\ref{eq:x_mix}). With the training distribution enlarged, the model generalization gets improved.

Moreover, we incorporate entropy minimization for the mixed data as well:
\begin{align}
    L_{ent}' = & -\frac{1}{n} \sum_{\bm{x_{1}},\bm{x_{2}} \in \puset} l_{bce}(s',s'). \label{eq:ent_}
\end{align}

Finally, we present the overall objective of the label distribution alignment with entropy minimization and Mixup, denoted as \textbf{Dist-PU}, as follows:
\begin{equation}
    L_{dist} = \hat{R}_{lab} + \mu L_{ent} + \nu L_{mix} + \gamma L_{ent}', \label{eq:final_obj}
\end{equation}
where $\mu$ ajusts the importance of $L_{ent}$; $\nu$ and $\gamma$ control the strength of $L_{mix}$ and $L_{ent}'$, respectively.

\section{Experiment}

\begin{table*}[htp]\small
  \centering
  \caption{Summary of used datasets and their corresponding models.}
  \setlength{\tabcolsep}{2.7mm}{
    \begin{tabular}{cccccccc}
    \hline
    Dataset & Input Size & $n_{L}$ & $n_{U}$ & \# Testing &    $\pi_{P}$   & Positive Class & Model \\
    \hline
    F-MNIST & $28 \times 28$ & 500   & 60,000 & 10,000 & 0.4   & Top (\ie, 0, 2, 4 and 6) & 6-layer MLP \\
    CIFAR-10 & $3 \times 32 \times 32$ & 1,000 & 50,000 & 10,000 & 0.4   & Vehicles (\ie, 0, 1, 8 and 9) & 13-layer CNN \\
    Alzheimer & $3 \times 224 \times 224$ & 769 & 5,121 & 1,279 & 0.5   & Alzheimer's Disease & ResNet-50 \cite{he2016deep} \\
    \hline
    \end{tabular}%
    }
  \label{tab:dataset}%
\end{table*}%
\begin{table*}[htp]\small
  \centering
  \caption{Comparative results on F-MNIST, CIFAR-10, and Alzheimer. \best{Best} and \secondb{second best} values are both highlighted. \ccmark (\xxmark) denotes \\ that our Dist-PU is significantly better (worse) than the corresponding methods revealed by the paired t-test with confidence level 95\%.}
  \setlength{\tabcolsep}{2.2mm}{
    \begin{tabular}{c|c|cccccc}
    \hline
    Dataset & Method  & ACC (\%) & Prec. (\%) & Rec. (\%) & F1 (\%) & AUC (\%) & AP (\%) \\
    \hline
    \hline
    \multicolumn{1}{c|}{\multirow{11}[4]{*}{F-MNIST}} & naive & 91.07 (0.93) \ccmark & 90.16 (2.25) \ccmark & 87.28 (2.08) \ccmark & 88.66 (1.14) \ccmark & 96.94 (0.67) \ccmark & 94.27 (1.40) \ccmark \\
          & uPU   & 94.02 (0.30) \ccmark & 92.50 (1.26) \ccmark & 92.59 (0.80) \ccmark & 92.53 (0.31) \ccmark & 97.34 (0.54) \ccmark & 96.60 (0.52) \ccmark \\
          & nnPU  & 94.44 (0.49) \ccmark & 91.69 (1.13) \ccmark & \secondb{94.69 (0.84) \blankM} & 93.16 (0.57) \ccmark & 97.53 (0.48) \ccmark & 96.39 (0.98) \ccmark \\
          & RP    & 92.37 (1.08) \ccmark & 88.58 (1.56) \ccmark & 92.94 (2.38) \ccmark & 90.69 (1.39) \ccmark & 97.14 (0.58) \ccmark & 94.39 (1.31) \ccmark \\
          & PUSB  & 94.50 (0.36) \ccmark & \secondb{93.12 (0.44) \ccmark} & 93.12 (0.44) \ccmark & 93.12 (0.44) \ccmark & 97.31 (0.50) \ccmark & 96.28 (0.96) \ccmark \\
          & PUbN  & \secondb{94.82 (0.16)} \ccmark & 92.92 (0.50) \ccmark & 94.24 (0.93) \blankM & \secondb{93.57 (0.24) \ccmark} & \secondb{98.33 (0.13) \blankM } & 96.72 (0.23) \ccmark \\
          & Self-PU & 94.75 (0.25) \ccmark & 91.73 (0.80) \ccmark & \best{95.50 (0.61) \blankM} & 93.57 (0.28) \ccmark & 97.62 (0.31) \ccmark &  96.14 (0.70) \ccmark \\
          & aPU & 94.71 (0.34) \ccmark & 92.71 (0.50) \ccmark & 94.20 (1.06) \blankM & 93.44 (0.45) \ccmark & 97.67 (0.40) \ccmark & 96.64 (0.48) \ccmark \\
          & VPU   & 92.26 (1.11) \ccmark & 89.04 (2.00) \ccmark & 92.01 (2.00) \ccmark & 90.48 (1.35) \ccmark & 97.38 (0.44) \ccmark & 95.57 (0.62) \ccmark \\
          & ImbPU & 94.54 (0.42) \ccmark & 92.81 (1.53) \ccmark & 93.66 (1.67) \blankM & 93.21 (0.52) \ccmark & 97.67 (0.81) \ccmark & \secondb{96.73 (0.86) \ccmark} \\
\cline{2-8}          & Dist-PU & \best{95.40 (0.34) \blankM} & \best{94.18 (0.90) \blankM} & 94.34 (1.00) \blankM & \best{94.25 (0.43) \blankM} & \best{98.57 (0.24) \blankM} & \best{97.90 (0.30) \blankM} \\
    \hline
    \hline
    \multirow{11}[3]{*}{CIFAR-10} & naive & 84.92 (0.89) \ccmark & 83.59 (0.89) \ccmark & 77.57 (3.23) \ccmark & 80.43 (1.56) \ccmark & 92.29 (0.63) \ccmark & 88.23 (0.79) \ccmark \\
          & uPU   & 88.35 (0.45) \ccmark & 87.18 (2.39) \ccmark & 83.23 (2.68) \ccmark & 85.10 (0.56) \ccmark & 94.91 (0.62) \ccmark & 92.62 (1.11) \ccmark \\
          & nnPU  & 88.89 (0.45) \ccmark & 86.18 (1.15) \ccmark & 86.05 (1.42) \ccmark & 86.10 (0.58) \ccmark & 95.12 (0.52) \ccmark & 92.42 (1.38) \ccmark \\
          & RP    & 88.73 (0.15) \ccmark & 86.01 (1.01) \ccmark & 85.82 (1.51) \ccmark & 85.90 (0.32) \ccmark & 95.17 (0.23) \ccmark & 92.92 (0.56) \ccmark \\
          & PUSB  & 88.95 (0.41) \ccmark & 86.19 (0.51) \ccmark & 86.19 (0.51) \ccmark & 86.19 (0.51) \ccmark & 95.13 (0.52) \ccmark & 92.44 (1.34) \ccmark \\
          & PUbN  & \secondb{89.83 (0.30) \ccmark} & \secondb{87.85 (0.98) \ccmark} & 86.56 (1.87) \ccmark & \secondb{87.18 (0.54) \ccmark} & 94.44 (0.35) \ccmark & 91.28 (1.11) \ccmark \\
          & Self-PU & 89.28 (0.72) \ccmark & 86.16 (0.78) \ccmark & \secondb{87.21 (2.35) \ccmark} & 86.67 (1.06) \ccmark & 95.47 (0.58) \ccmark & 93.28 (1.01) \ccmark \\
          & aPU & 89.05 (0.52) \ccmark & 86.29 (1.30) \ccmark & 86.37 (0.79) \ccmark & 86.32 (0.56) \ccmark & 95.09 (0.42) \ccmark & 92.41 (1.23) \ccmark \\
          & VPU   & 87.99 (0.48) \ccmark & 86.72 (1.41) \ccmark & 82.71 (2.84) \ccmark & 84.63 (0.91) \ccmark & 94.51 (0.41) \ccmark & 92.00 (0.73) \ccmark \\
          & ImbPU & 89.41 (0.46) \ccmark & 86.69 (0.87) \ccmark & 86.87 (0.82) \ccmark & 86.77 (0.56) \ccmark & \secondb{95.52 (0.27) \ccmark} & \secondb{93.45 (0.45) \ccmark} \\
\cline{2-8}          & Dist-PU & \best{91.88 (0.52) \blankM} & \best{89.87 (1.09) \blankM} & \best{89.84 (0.81) \blankM} & \best{89.85 (0.62) \blankM} & \best{96.92 (0.45) \blankM} & \best{95.49 (0.72) \blankM} \\
\hline
\hline
    \multirow{11}[3]{*}{Alzheimer} & naive & 61.45 (3.81) \ccmark & 62.51 (5.87) \ccmark & 61.44 (12.5) \ccmark & 61.05 (4.65) \ccmark & 66.26 (6.39) \ccmark & 63.28 (5.98) \ccmark \\
          & uPU   & 68.48 (2.15) \ccmark & \best{69.65 (3.50)} \blankM & 66.13 (6.13) \ccmark & 67.62 (2.78) \ccmark & 73.75 (2.94) \ccmark & 69.53 (3.23) \ccmark \\
          & nnPU  & 68.33 (2.13) \ccmark & 68.01 (2.33) \blankM & 69.48 (7.15) \ccmark & 68.55 (3.16) \ccmark & 72.90 (2.80) \ccmark & 69.45 (2.87) \ccmark \\
          & RP    & 61.61 (3.20) \ccmark & 61.89 (4.54) \ccmark & 64.60 (15.89) \blankM & 62.10 (5.61) \ccmark & 66.13 (3.28) \ccmark & 63.82 (2.29) \ccmark \\
          & PUSB  & 69.21 (2.39) \blankM & 69.16 (2.39) \blankM  & 69.26 (2.39) \ccmark & 69.21 (2.39) \ccmark & 74.43 (2.41) \ccmark & 70.00 (1.56) \ccmark \\
          & PUbN & 69.98 (1.34) \ccmark & \secondb{69.42 (2.49) \blankM} & 72.02 (8.42) \ccmark & 70.38 (3.19) \blankM & 74.97 (1.01) \ccmark & 69.89 (1.47) \ccmark \\
          & Self-PU & \secondb{70.88 (0.72) \ccmark} & 69.32 (2.54) \blankM & 75.43 (5.07) \blankM & \secondb{72.09 (1.09) \ccmark} & \secondb{75.89 (1.79) \ccmark} & \secondb{71.68 (3.84) \blankM} \\
          & aPU & 68.52 (1.75) \ccmark & 66.21 (0.91) \ccmark & 75.71 (8.20) \blankM & 70.46 (3.35) \blankM & 73.80 (2.59) \ccmark & 70.71 (3.70) \ccmark \\
          & VPU   & 67.44 (0.65) \ccmark & 64.74 (1.12) \ccmark & \secondb{76.68 (3.60) \blankM} & 70.16 (1.08) \ccmark & 73.12 (0.85) \ccmark & 71.11 (0.75) \blankM \\
          & ImbPU & 68.18 (0.83) \ccmark & 67.54 (2.52) \ccmark & 70.64 (6.54) \ccmark & 68.83 (1.94) \ccmark & 73.81 (0.71) \ccmark & 70.46 (1.07) \ccmark \\
\cline{2-8}          & Dist-PU & \best{71.57 (0.62) \blankM} & 68.48 (1.16) \blankM & \best{80.09 (5.10) \blankM} & \best{73.74 (1.64) \blankM} & \best{77.13 (0.69) \blankM} & \best{73.33 (1.47) \blankM} \\
    \hline
    \hline
    \end{tabular}%
    }
  \label{tab:com_results}%
\end{table*}%

\subsection{Experimental settings}
\textbf{Datasets.} We conduct experiments on three benchmarks: F-MNIST \cite{xiao2017/online} for fashion product classification, CIFAR-10 \cite{krizhevsky2009learning} for vehicle class identification, together with the Alzheimer dataset\footnote{Dubey, S. Alzheimer’s Dataset. Available online: \\ https://www.kaggle.com/tourist55/alzheimers-dataset-4-class-of-images} for the recognition of the Alzheimer’s Disease. More details are concluded in Tab.\ref{tab:dataset}.

\textbf{Evaluation metrics.} \bangyan{For each model, we report six metrics on the test set for a more comprehensive comparison, including} accuracy (ACC), Precision (Prec.), Recall (Rec.), F1, Area Under ROC Curve (AUC) and Average Precision (AP). \bangyan{Experiments are repeated with five random seeds, then the mean and the standard deviation of each metric are recorded.} 
\begin{table*}[htp]\small
  \centering
  \caption{Ablation results on CIFAR-10 with \cmark indicating the enabling of the corresponding loss term.}
  \setlength{\tabcolsep}{1.8mm}{
    \begin{tabular}{cccccccccc}
    \hline
    Variant & $\hat{R}_{lab}$ & $L_{ent}$ / $L_{ent}'$ & $L_{mix}$ & ACC (\%) & Prec. (\%) & Rec. (\%) & F1 (\%) & AUC (\%) & AP (\%) \\
    \hline
    I     & \xmark     & \xmark     & \xmark     & 84.92 (0.89) & 83.59 (0.89) & 77.57 (3.23) & 80.43 (1.56) & 92.29 (0.63) & 88.23 (0.79) \\
    II    & \cmark     & \xmark     & \xmark     & 89.30 (0.48) & 86.96 (0.86) & 86.19 (1.12) & 86.57 (0.52) & 95.40 (0.52) & 93.22 (0.97) \\
    III   & \xmark     & \cmark     & \xmark     & 87.50 (0.24) & 83.55 (0.17) & 85.62 (0.78) & 84.57 (0.36) & 93.61 (0.33) & 89.36 (0.98) \\
    IV    & \xmark     & \xmark     & \cmark     & 86.97 (0.23) & 84.50 (4.35) & 83.22 (6.47) & 83.57 (1.11) & 94.37 (0.22) & 91.21 (0.73) \\
    V     & \cmark     & \cmark     & \xmark     & 89.41 (0.48) & 86.40 (0.98) & 87.30 (1.75) & 86.83 (0.69) & 95.61 (0.44) & 93.51 (0.46) \\
    VI    & \cmark     & \xmark     & \cmark     & \secondb{91.62 (0.47)} & \best{90.96 (1.07)} & \secondb{87.78 (0.32)} & \secondb{89.34 (0.54)} & \secondb{96.80 (0.44)} & \best{95.54 (0.35)} \\
    VII   & \xmark     & \cmark     & \cmark     & 87.67 (0.52) & 85.01 (1.78) & 84.07 (1.90) & 84.51 (0.60) & 94.32 (0.35) & 90.65 (0.80) \\
    \hline
    VIII  & \cmark     & \cmark     & \cmark     & \best{91.88 (0.52)} & \secondb{89.87 (1.09)} & \best{89.84 (0.81)} & \best{89.85 (0.62)} & \best{96.92 (0.45)} & \secondb{95.49 (0.72)} \\
    \hline
    \end{tabular}%
    }
  \label{tab:ablation}%
\end{table*}

\textbf{Implementation details.} All the experiments are run on Geoforce RTX 3090 \bangyan{implemented by PyTorch}. Backbones of each dataset are summarized in Tab.\ref{tab:dataset}. For CIFAR-10, we only normalize the input images with mean = (0.485, 0.456, 0.406) and std = (0.229, 0.224, 0.225) \bangyan{without any extra data preprocessing techniques}. Besides, we clamp the logits between $-10$ and $10$ to avoid the potantial NaN error from Eq.~(\ref{eq:ent},\ref{eq:ent_}). The training batch size is set as 256 for F-MNIST and CIFAR-10, while 128 for Alzheimer. We use Adam \cite{DBLP:journals/corr/KingmaB14} as the optimizer with a cosine annealing scheduler \cite{DBLP:conf/iclr/LoshchilovH17}, where the initial learning rate is set as $5\times10^{-4}$; weight decay is set as $5\times 10^{-3}$. Dist-PU first experiences a warm-up stage of several epochs without Mixup, then trains with Mixup for another 60 epochs, where $\mu$ is also with a cosine annealing scheduling to mitigate overfitting. Moreover, the values of hyperparameters $\mu, \nu, \gamma$ and $\alpha$ are searched within the range of $[0, 0.1], [0, 10], [0, 0.3]$ and $[0.1, 10]$, respectively.

\subsection{Comparision with state-of-the-art methods}

\par{\textbf{Competitors.}} We compare our Dist-PU with a naive baseline which constructs the negative class with randomly sampled unlabeled data, together with $9$ competitive PU algorithms including uPU \cite{DBLP:conf/nips/PlessisNS14}, nnPU \cite{DBLP:conf/nips/KiryoNPS17}, {RP}  \cite{northcutt2017rankpruning},  {PUSB}  \cite{DBLP:conf/iclr/KatoTH19},  {PUbN}  \cite{DBLP:conf/icml/HsiehNS19},  {Self-PU}  \cite{DBLP:conf/icml/ChenCCYG0W20},  {aPU} \cite{DBLP:conf/nips/HammoudehL20},  {VPU}  \cite{DBLP:conf/nips/ChenLWZW20}, and  {ImbPU}  \cite{DBLP:conf/ijcai/SuCX21}. Due to the space limitation, the detailed descriptions are provided in supplementary materials. 

\textbf{Results.} \bangyan{The results on all the datasets are recorded} in Tab.\ref{tab:com_results}. \bangyan{It shows that our proposed} Dist-PU outperforms the competitors by a significant margin for all the datasets \bangyan{in terms of most metrics}, surpassing the second-best roughly by 1\% to 3\% on average. This validates the \bangyan{effectiveness} of our proposed method. Moreover, our Dist-PU also achieves a relative balance between precision and recall metrics. Besides, some observations could be made: (1) We see that the performance of classic cost-sensitive PU approaches including uPU, nnPU, aPU and ImbPU fluctuate over different datasets. This might be because their objectives consist of the loss classifying unlabeled data to the negative class, thus they may suffer from overfitting and tend to make negative predictions. On the contrary, our Dist-PU alleviates this problem via the proposed label distribution alignment. (2) Self-PU and PUbN are the most competitive baselines. Their success can be partly attributed to some extra designs. For example, Self-PU adopts mentor net and self-paced learning, while PUbN relies on the pre-trained model by nnPU. In the ablation study, we will show that our method still performs well without Mixup regularization.
(3) Some competitors such as VPU, performs relatively less promising. 
Notably, considering that VPU does not use the class prior information, it might not be able to outperform other baselines given the accurate class prior. Nevertheless, sensitivity analysis in supplementary materials shows that our Dist-PU is relatively robust to misspecified prior.

\begin{figure}[t]
    \centering
    \subfigure[w/o Mixup]{
    \centering
    \begin{minipage}[t]{0.225\textwidth}
        \includegraphics[width=1.5in]{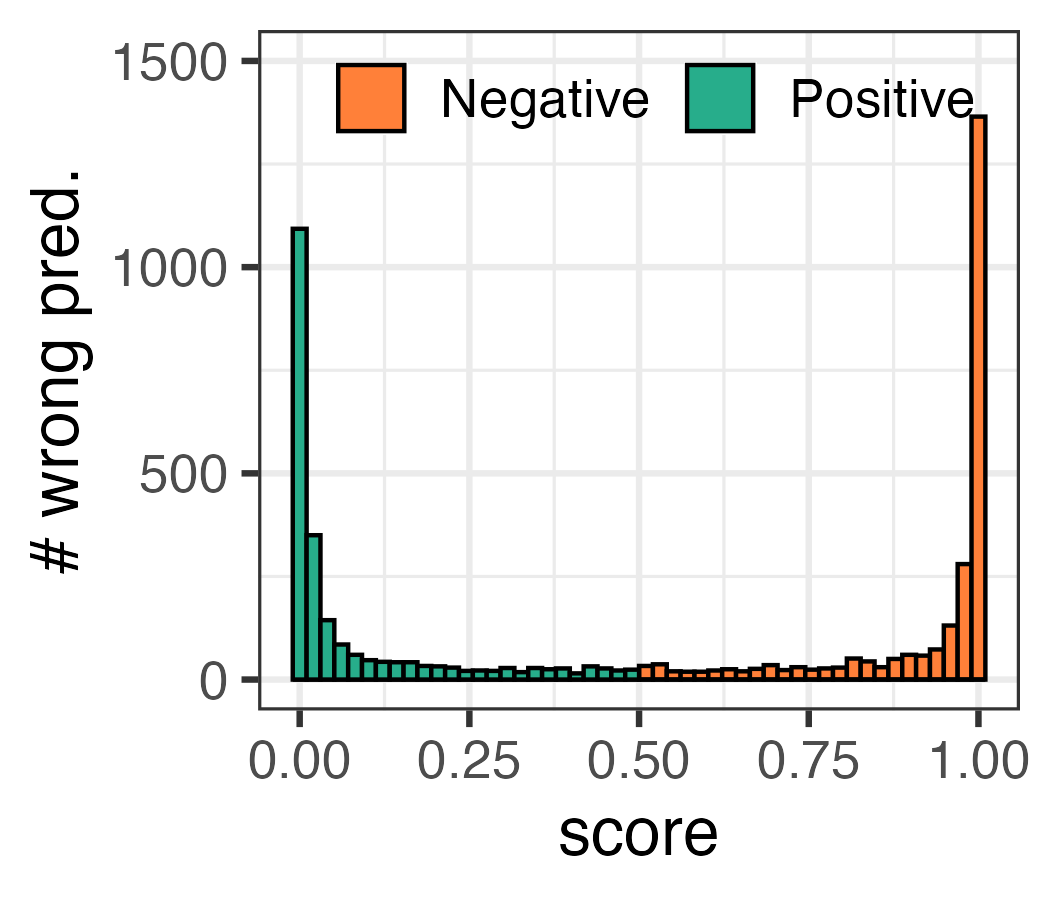}
    \end{minipage}
    }
    \subfigure[w/ Mixup]{
    \centering
    \begin{minipage}[t]{0.225\textwidth}
        \includegraphics[width=1.5in]{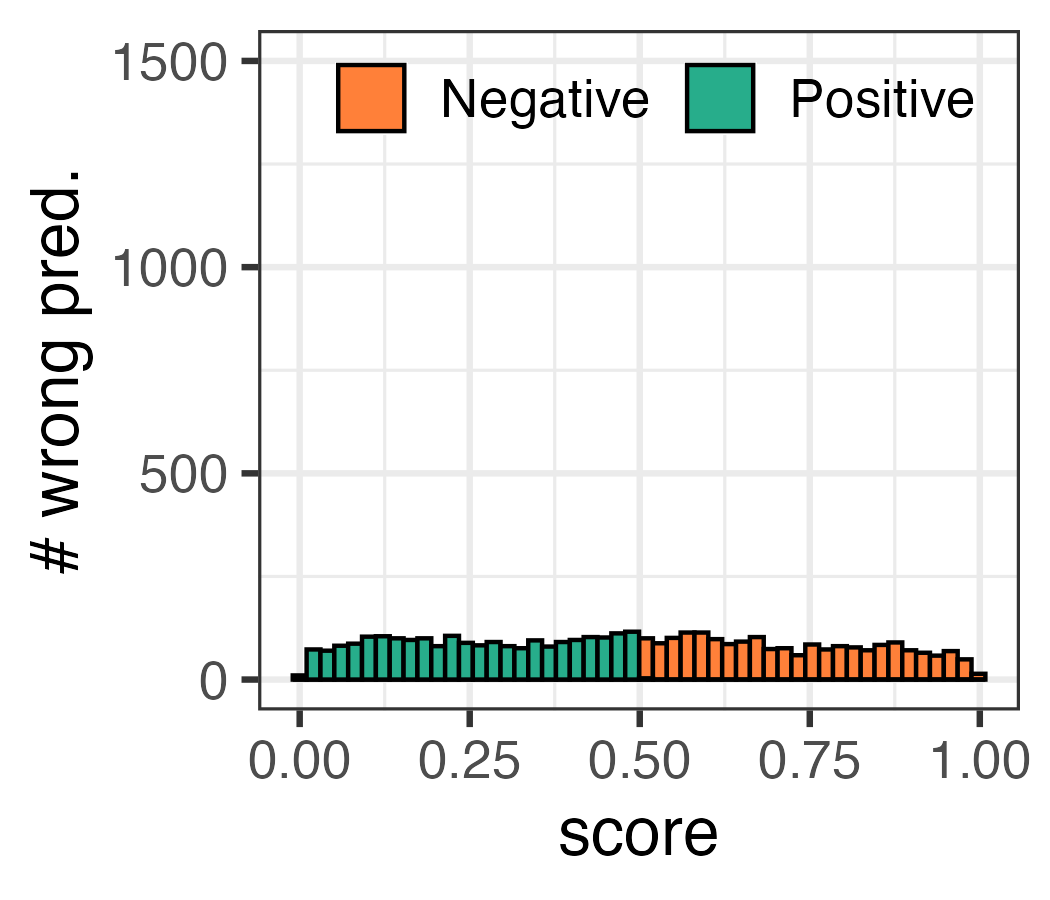}
    \end{minipage}
    }
    \caption{Histogram of wrongly-predicted scores ($s$) from the training set on CIFAR-10.}
    \label{fig:hist_mixup}
\end{figure}

\subsection{Ablation studies}
In order to analyze the impact of \bangyan{each module in Dist-PU, \ie,} label distribution alignment, entropy minimization, and Mixup regurlarization, we conduct ablation studies on CIFAR-10. The results are shown in Tab.\ref{tab:ablation}. \yunrui{From the table, we could get the following observations: (1) Our label distribution alignment plays an essential role in Dist-PU. By comparing the results of the variants I w/ II, IV w/ VI, and VII w/ VIII, we can find that our label distribution alignment leads to a significant improvement from those without label distribution alignment. (2) Entropy minimization exhibits a minor enhancement, which can be observed from comparisons of variants II w/ V and VI w/ VIII. (3) Mixup contributes to the performance especially when label distribution alignment is incorporated. As shown by variants III and VII, Mixup brings few gains for entropy minimization without label distribution alignment. However, with label distribution alignment, the increase from Mixup is much more noticeable based on comparisons of variants II with VI and V with VIII. This also reflects another advantage of label distribution alignment in a way that it promotes the effect of Mixup.}

For better understanding the \bangyan{impact} of Mixup on alleviating confirmation bias, we plot the scores of wrong predictions on the training data of CIFAR-10 in Fig.~\ref{fig:hist_mixup}. As the result shows, with Mixup, the total number of wrong predictions becomes less. Besides, the score distribution is more smooth with Mixup, which means the confidence of wrong predictions is lower in general. This unveils some intrinsic mechanisms of Mixup in reducing confirmation bias.


\subsection{Effectiveness of hyper-parameters}

\begin{figure}[t]
    \centering
    \subfigure{
    \centering
    \begin{minipage}[t]{0.225\textwidth}
        \includegraphics[width=1.5in]{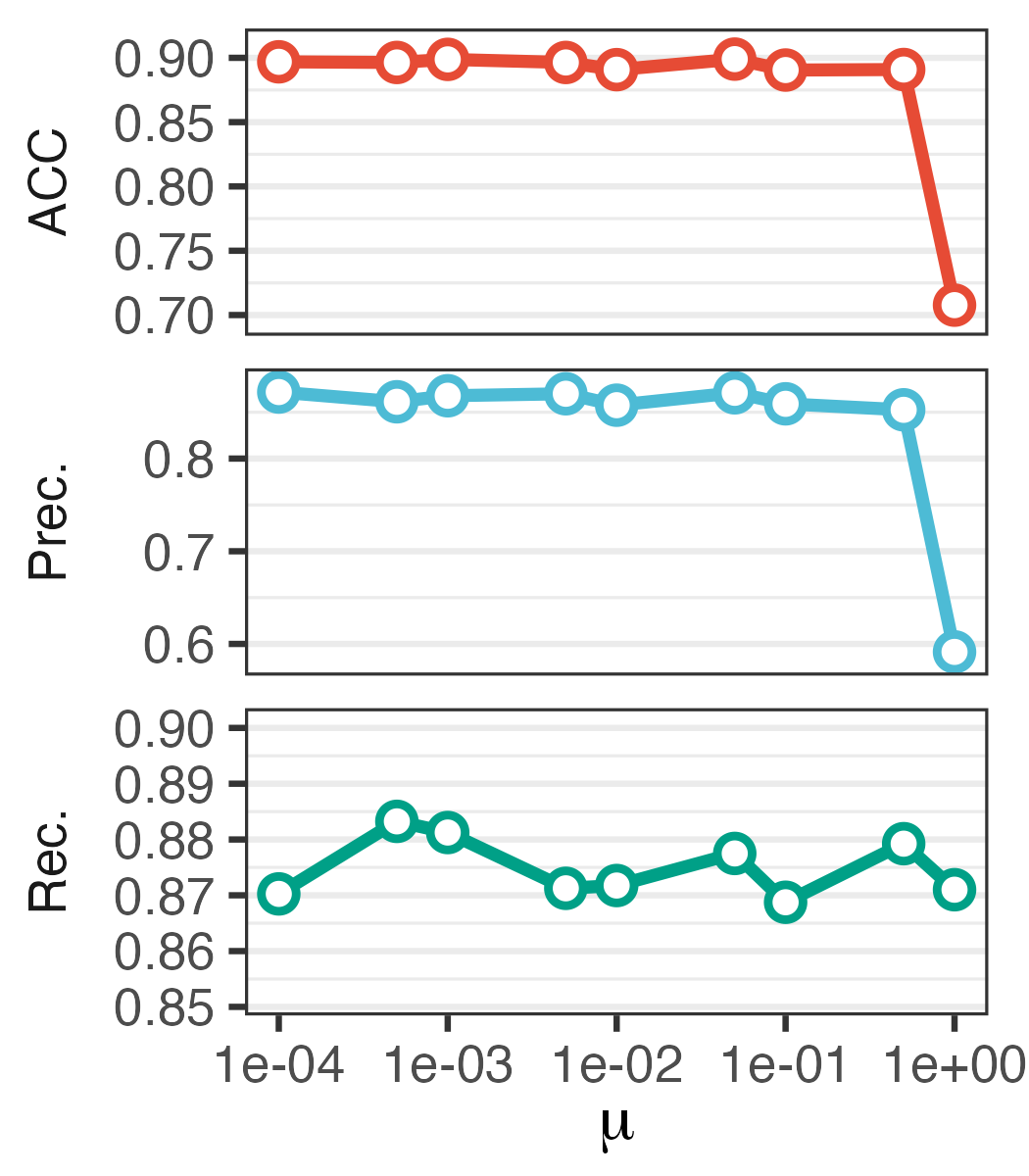}
    \end{minipage}
    }
    \subfigure{
    \centering
    \begin{minipage}[t]{0.225\textwidth}
        \includegraphics[width=1.5in]{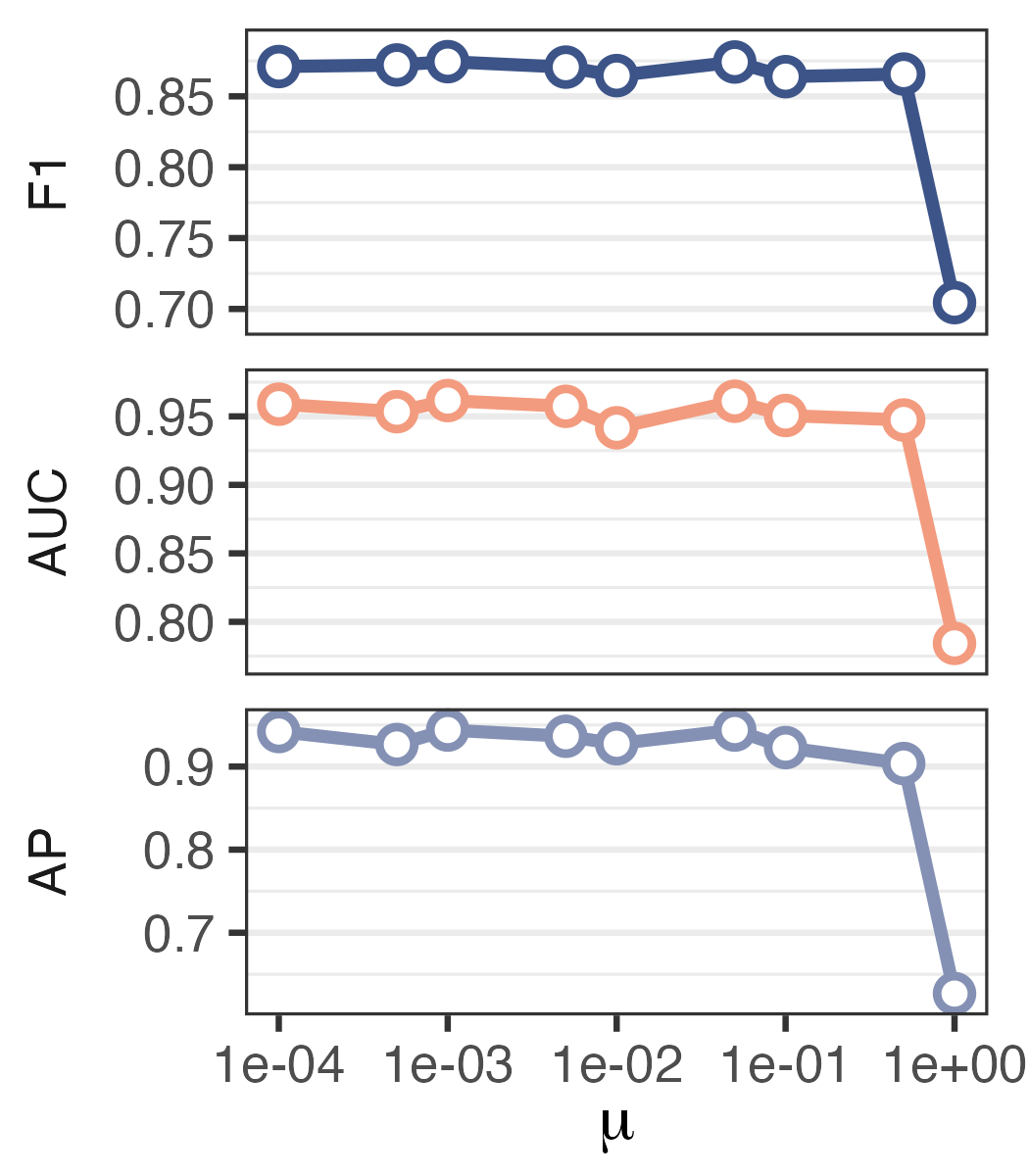}
    \end{minipage}
    }
    \caption{Influences of different $\mu$ on CIFAR-10.}
    \label{fig:line_mu}
\end{figure}

\textbf{Effectiveness of $\mu$ for entropy minimization on $\xdistu$.} Parameter $\mu$ controls the concentration degree of the predicted score distribution of the unlabeled data. To get rid of the interference of other loss terms, we conduct experiments with different $\mu$ without Mixup. In Fig.~\ref{fig:line_mu}, we can see that when $\mu$ locates between $0$ and $0.1$, all the metrics except Rec. \bangyan{are relatively stable}. However, a $\mu$ larger than $0.5$ heavily \bangyan{degenerates} the model performance. 

\begin{figure}[t]
    \centering
    \subfigure{
    \centering
    \begin{minipage}[t]{0.225\textwidth}
        \includegraphics[width=1.5in]{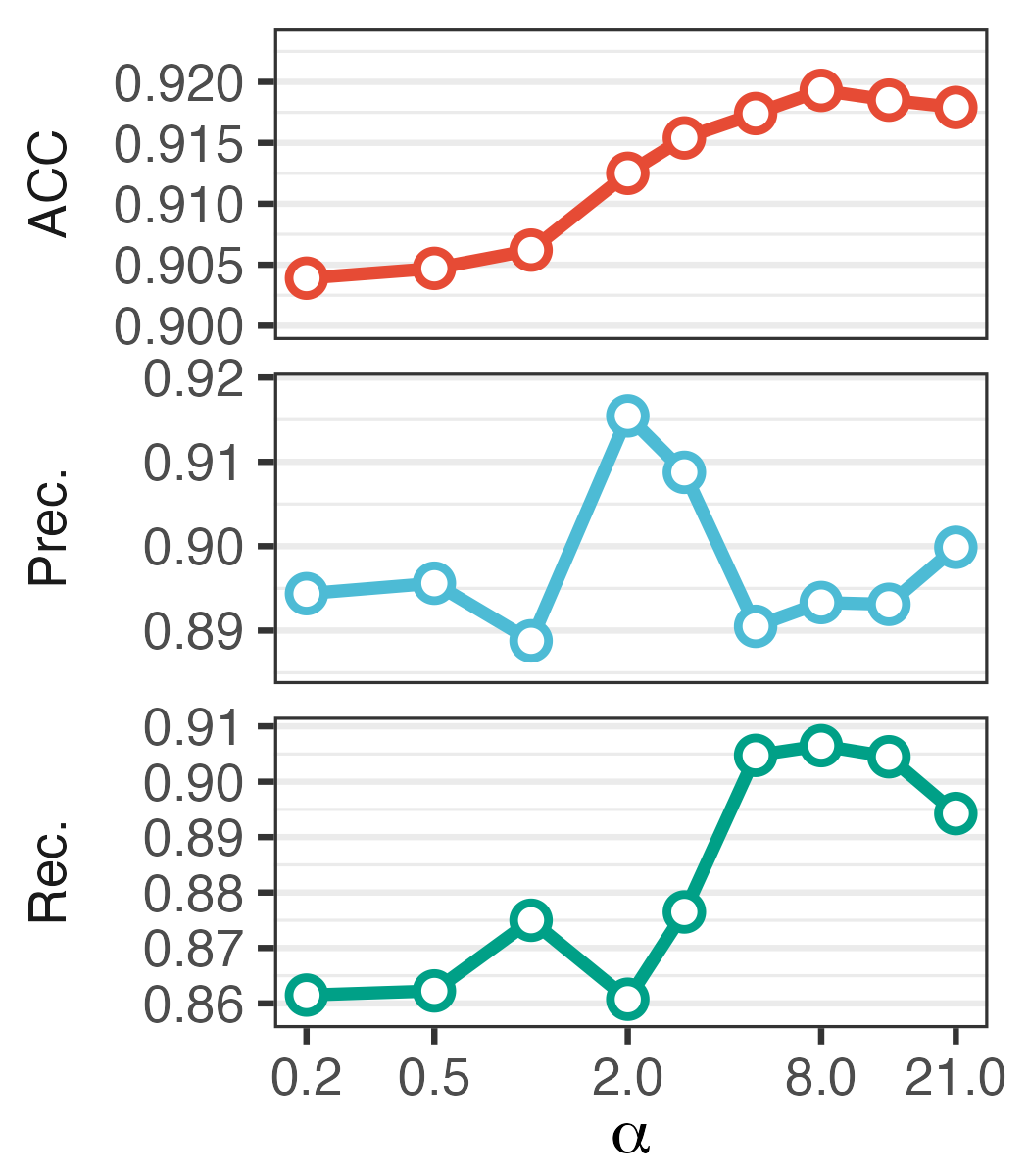}
    \end{minipage}
    }
    \subfigure{
    \centering
    \begin{minipage}[t]{0.225\textwidth}
        \includegraphics[width=1.5in]{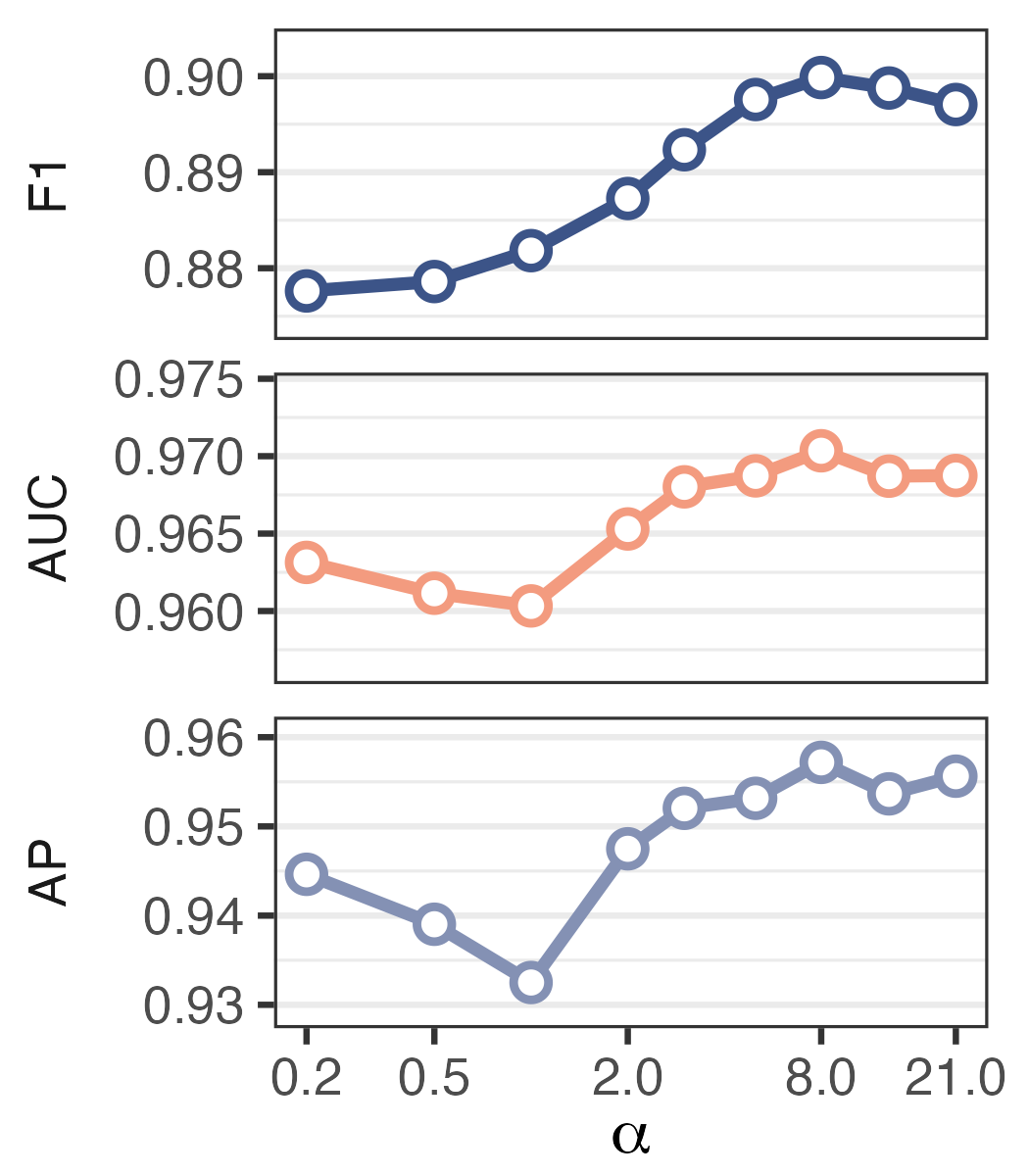}
    \end{minipage}
    }
    \caption{Influences of different $\alpha$ on CIFAR-10.}
    \label{fig:line_alpha}
\end{figure}

\begin{figure}
    \centering
    \includegraphics[width=0.4\textwidth]{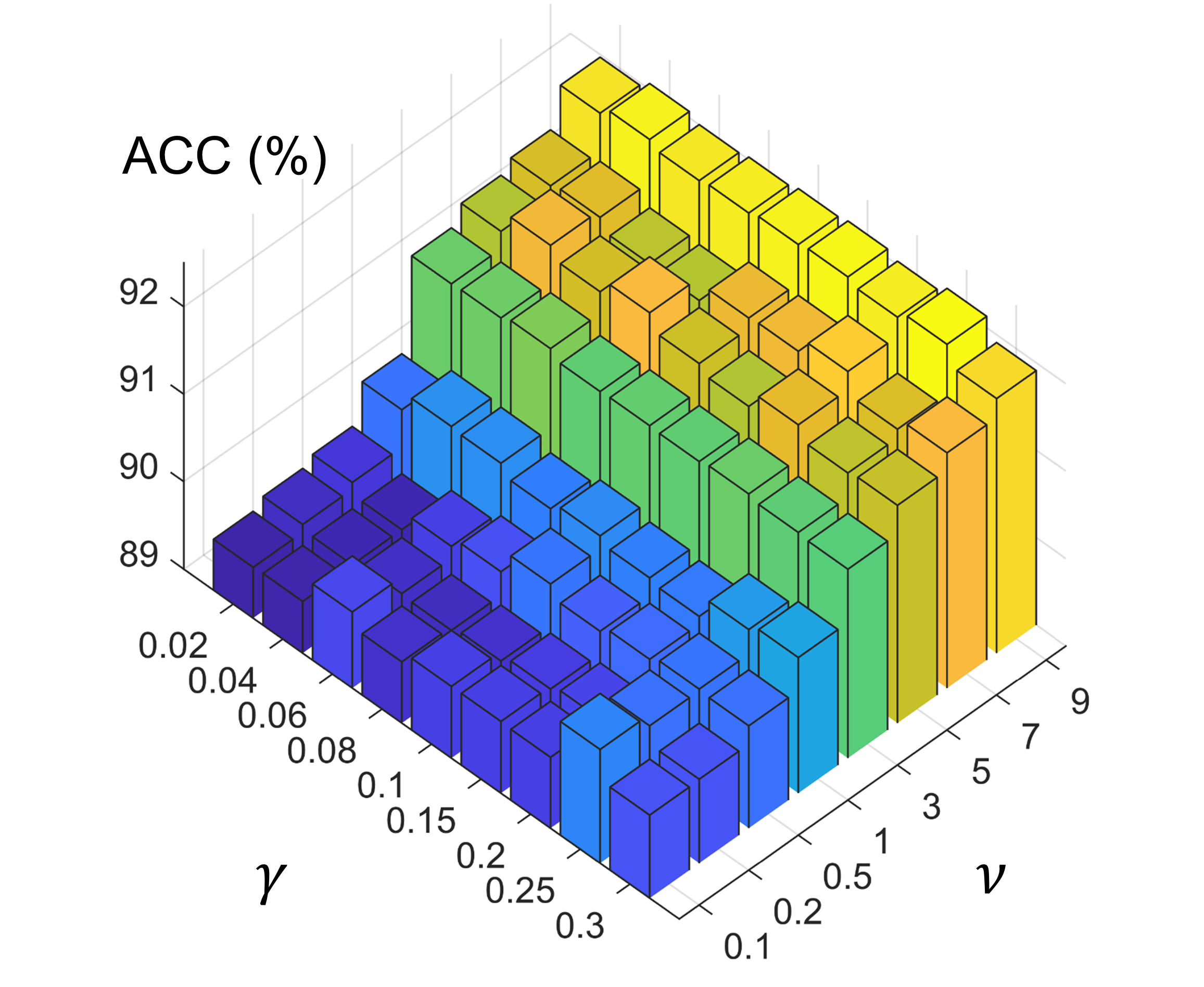}
    \caption{Influences of different $\nu$ and $\gamma$ on CIFAR-10.}
    \label{fig:heat_map}
\end{figure}

\textbf{Effectiveness of $\alpha$ for Beta distribution in Mixup.} Parameter $\alpha$ controls the shape of Beta distribution in Mixup. The higher $\alpha$ is, the narrower its curve will be and $\lambda$ will be more likely near $0.5$. Fig.~\ref{fig:line_alpha} shows that rationally larger $\alpha$ helps improve the performance. Understandably, a larger $\alpha$ means more chance to get a $\lambda$ near $0.5$ and a better-mixed target for the mixed data, especially when its source labels are different, thus better reducing confirmation bias.

\textbf{Effectiveness of $\nu$ and $\gamma$ for Mixup.} Parameter $\nu$ controls the power of Mixup on alleviating confirmation bias. In Fig.~\ref{fig:heat_map}, we see that the larger $\nu$ is, the better confirmation bias is weakened and the higher ACC our algorithm could reach. It also shows that a moderately larger $\gamma$, which leads to lower entropy of the mixed predictions, has positive effects on the performance.


\section{Conclusion}
This paper proposes \bangyan{a novel PU learning method from a label distribution alignment perspective called \textbf{Dist-PU}}. Specifically, Dist-PU pursues the label distribution consistency between the predictions and the ground-truth labels over the labeled positive and unlabeled data. It then leverages entropy minimization to make the positive-negative distributions more separable. With Mixup to mitigate the confirmation bias, Dist-PU consistently outperforms the state-of-the-art methods over most metrics on real-world datasets including F-MNIST, CIFAR-10 and Alzheimer. Ablation studies and sensitivity analysis further demonstrate the effectiveness of each module in Dist-PU. We hope that the proposed label distribution alignment scheme could provide some enlightenment on other weakly supervised scenarios as well, especially for those with unlabeled or inaccurately labeled data but of known label distributions.

\section*{Acknowledgments}
This work was supported in part by the National Key R\&D Program of China under Grant 2018AAA0102000, in part by National Natural Science Foundation of China: U21B2038, 61931008, 61836002, 6212200758 and 61976202, in part by the Fundamental Research Funds for the Central Universities, in part by Youth Innovation Promotion Association CAS, in part by the Strategic Priority Research Program of Chinese Academy of Sciences, Grant No. XDB28000000.

{\small
\bibliographystyle{ieee_fullname}
\bibliography{egbib}
}

\end{document}